\def\year{2017}\relax
\documentclass[letterpaper]{article}
\usepackage{aaai17}
\usepackage{times}
\usepackage{helvet}
\usepackage{courier}

\usepackage[T1]{fontenc}    
\usepackage{url}            
\usepackage{booktabs}       
\usepackage{amsfonts}       
\usepackage{nicefrac}       
\usepackage{microtype}      
\usepackage{amssymb}
\usepackage{multirow}
\usepackage{times}
\usepackage{amsmath}
\usepackage{amsthm}
\usepackage{array}
\usepackage{bm}
\usepackage[]{algorithm2e}
\usepackage{balance}
\usepackage{graphicx}
\usepackage{subfigure}
\usepackage[square,sort]{natbib}
\usepackage[toc,page]{appendix}
\newcommand{\comment}[1]{}
\usepackage{float}

\begin{document}
\title{Tensor Decomposition via Variational Auto-Encoder}

\author{Bin Liu, Zenglin Xu \\ School of Computer Science \& Engineering \\ University of Electronic Science and Technology of China \\ 2006 Xiyuan Avenue, West Hi-tech Zone, Chengdu, China \\ liu@std.uestc.edu.cn, zlxu@uestc.edu.cn
\And
Yingming Li \\College of Electronic Engineering\\  Zhejiang University \\ 38 Zheda Road, Hangzhou, China \\ liym006@163.com}

\maketitle
\begin{abstract}
Tensor decomposition is an important technique for capturing the high-order interactions among multiway data. Multi-linear tensor composition methods, such as the Tucker decomposition and the CANDECOMP/PARAFAC (CP), assume that the complex interactions among objects are multi-linear, and are thus insufficient to represent nonlinear relationships in data. Another assumption of these methods is that a predefined rank should be known. However, the rank of tensors is hard to estimate, especially for cases with missing values. To address these issues, we design a Bayesian generative model for tensor decomposition. Different from the traditional Bayesian methods,  the high-order interactions of tensor entries are modeled with variational auto-encoder. The proposed model takes advantages of Neural Networks and  nonparametric Bayesian models, by replacing the multi-linear product in traditional Bayesian tensor decomposition with a complex nonlinear function (via Neural Networks) whose parameters can be learned from data. Experimental results on synthetic data and real-world chemometrics tensor data have demonstrated that our new model can achieve significantly higher prediction performance than the state-of-the-art tensor decomposition approaches.
\end{abstract}

\section{Introduction}
Multiway arrays (i.e., tensors) are very popular in data analysis in a number of domains. Tensors can represent the high-order interactions of data effectively and faithfully. For example, in bioinformatics, a <patient, medicine, biomarker> tensor can represent the test effects of the $i$-th patient on the $k$-th  medicine with respective to the $j$-th biomarker. As an effective way to tensor analysis, tensor decomposition can analyze such a high-order tensor via its low-dimension embedding. A variety of algorithms for tensor factorization have been well studied in the past decades. 
Classical tensor decomposition methods include the Tucker decomposition \citep{tucker1966some}, the CANDE-COMP/PARAFAC (CP) \citep{harshman1970foundations} and their variants.
Especially, CP can be taken as a special case of Tucker decomposition, by constraining the core tensor of the Tucker model to be diagonal. The CP decomposition for a $D$-way tensor can represent the $D$-order correlations among data via just $D$ factor matrices.
However, whatever Tucker or CP, they all assume that the interactions between the latent factors are multi-linear. Therefore, they are not powerful enough to capture the complex and nonlinear relationships in reality. Another point is  the multi-linear algorithms excessively depend on the appropriate rank determination. An incorrect rank will cause overfitting for factors estimating.

Recently, modeling the nonlinear interaction in tensor decomposition with the probabilistic model is more and more attractive \citep{pragarauskas2010temporal, morup2009automatic, xu2011infinite, rai2014scalable, zhao2015bayesian}.
To solve the insufficient ability to model complex interactions among tensor entities, the Infinite Tucker decomposition (InfTucker) \citep{xu2011infinite} proposed tensor-variate latent nonparametric Bayesian models by extending the Tucker model to an infinite feature space. In addition, the authors of \citep{rai2014scalable, rai2015scalable} and \citep{zhao2015bayesian} generalized the CP decomposition in a Bayesian way to model the intricate and nonlinear interactions in tensor data.

Despite the aforementioned advantages, in practice, traditional Bayesian latent variable models suffer from several disadvantages.
Firstly, the priors of latent variables should be chosen carefully. In modeling the relationship between latent and observed variables, the prior should be conjugated with posterior for the convenience of inference.
Secondly, traditional Bayesian variable selection usually have  expensive computational cost. The posterior distribution usually does not have a closed-form expression, so posterior optimization has to rely on sampling methods or variational inference, which could be time-consuming especially when the number of predictors is large.
Therefore, it is meaningful to find a more powerful way to represent the latent variable and to optimize the mode in a more efficient way.

In this paper, we put forward a novel Bayesian tensor decomposition method, named as Variational Auto-Encoder CP (VAECP).
Specifically, we think that the tensor entries can be generated by a very complex and intractable random process determined by some latent variables. And we also assume that these latent variables have the Gaussian priors. Then the tensor entries can be generated from a  distribution conditioned on the latent variables. To capture this complex generative process from the latent variables to the observed data, we model the conditional distribution with neural networks. And it is the generative network in the auto-encoder. The input of the Neural Networks are the latent variables, and output are the parameters of distribution for link function which can predict the value of tensor entries.
The proposed method combine Neural Networks and latent nonparametric Bayesian model together. Different from the traditional Bayesian tensor decomposition, it replace the inner product of latent factors with a complex nonlinear function (via neural networks) and avoid the trouble of having to make a conjugated posterior.

However, introducing neural networks into Bayesian latent variable model makes the posterior distribution very complicated and  will inconvenience the optimization. Thanks to the reparametrization trick proposed in \citep{KingmaW13}, the expectation of reconstructing the likelihood (conditional distribution) over posterior distribution of latent variable can be approximated. Furthermore, there are no global representations of latent variables that are shared by all data points.
Therefore, our Bayesian model can be inferred efficiently with stochastic optimization rather than a time-consuming computation of traditional way.
Our experimental results on synthetic and chemometrics datasets demonstrate that our new model achieved significantly higher prediction performance than the  state-of-the-art tensor decomposition approaches.

\section{Related Work}
Existing works of tensor decomposition within probabilistic frameworks can be grouped into the Tucker family and the CP family. And both of them have been applied to two kinds of applications, tensor completion \citep{chu2009probabilistic,xu2015bayesian} and hidden pattern detection \citep{zhe2015scalable}.
Chu et al. proposed a latent variable model for the Tucker factorization \citep{chu2009probabilistic}. Their work based on the assumption that the structural dependency of an incomplete multiway array can be learned from partially observed entries. By extending Gaussian process latent variable models \citep{titsias2010bayesian} to tensor-variate process, Xu et al. designed the InfTucker for missing values prediction on both binary and continuous tensor data \citep{xu2011infinite}. We compare with this method in our experiments. In \citep{zhe2015scalable}, the authors combined the Dirichlet process mixture prior with a tensor-variate Gaussian process to identify unknown latent clusters in tensor modes.

CP decomposition is the other popular framework of tensor analysis.
Yoshii et al. studied the CP from the technique of matrix analysis, extending non-negative matrix factorization to positive semi-definite tensor decomposition~\citep{yoshii2013infinite}.
In \citep{chi2012tensors}, another multi-linear CP decomposition algorithm was proposed to model sparse count data. But it cannot predict missing values in the tensor.
Additional information of tensor is valuable in predicting missing values.
The authors in \citep{rai2015leveraging} suggested a two-layer decomposition for CP which considering the scenario of tensor modes having attribute information. Probabilistic tensor approaches usually perform well in tensor completion. Existing Bayesian CP models assign the latent factors with appropriate priors and optimize the posterior with Bayesian inference~\citep{zhao2015bayesian, yoshii2013infinite, pragarauskas2010temporal, rai2014scalable}.
In \citep{zhao2015bayesian}, the authors proposed two algorithms FBCP and FBCP-MP by assigning the latent factors with the Gamma prior.
There are also some studies of analyzing the latent pattern within the framework of CP \citep{pragarauskas2010temporal, dunson2009nonparametric, rai2014scalable, Maehara2016cp}.
Dunson \& Xing in \citep{dunson2009nonparametric} developed a nonparametric Bayesian method for motif discovery, and they also generate their algorithm for modeling multivariate unordered categorical data.
And an improvement in spirit to \citep{dunson2009nonparametric} was presented in \citep{rai2014scalable}. To detect the latent pattern in the future, an expected CP decomposition was put forwarded in \citep{Maehara2016cp} for the cases that tensor can be given as the sum of multiple tensors (e.g., the time series tensor).
As we discussed before, CP corresponds to the Tucker decomposition model with the core tensor is diagonal. There are some works on modeling tensor decomposition in Bayesian way that can apply to both Tucker and CP based on automatic relevance determination (ARD)\citep{morup2009automatic, zhao2015bayesian}.

\section{Variational Auto-Encoder CP Decomposition}
\subsection{Bayesian Tensor Decomposition}
Before presenting the model, we introduce the notations. Overall this paper, We denote tensor with swash letter, the matrix as a capital letter. For vector, we use lowercase letters, however, we also refer the capital letter with an index as a vector. For example, for a matrix $A$, $A_{i:}$ refers to its $i$th row and $A_{: j}$ is the $j$th column of matrix $A$.

Let a $D$-way (or  $D$-mode) tensor denoted by $\mathcal{Y}$ and $\mathcal{Y}\in \mathbb{R}^{N_1\times N_2 \times \cdot\cdot\cdot\times N_D}$ with the number $N_d$ being the dimension of $\mathcal{Y}$ along the $d$-th way. We further suppose that it can be decomposed as $\mathcal{Y}=\mathcal{X}+\boldsymbol{\varepsilon}$, where $\mathcal{X}$ is a low-rank tensor and $\boldsymbol{\varepsilon}$ is a noise term and $\boldsymbol{\varepsilon}\sim \prod_{i_1,...,i_D}\mathcal{N}(0,\varepsilon^{-1})$.
Following \citep{harshman1970foundations, Kruskal1977Three,kolda2009tensor}, the CP decomposition decomposes a tensor  $\mathcal{X}$  into a sum of rank-1 component tensors as follows,
\begin{equation}\label{equ-cp}
\mathcal{X}=\sum_{r=1}^R U^{1}_{:r} \circ U^{2}_{:r} \circ \cdot \cdot \cdot \circ U^{D}_{:r} = [U^1,U^2,...,U^D],
\end{equation}
where $U^d=(U^{d}_{:1},...,U^{d}_{:r},...,U^{d}_{:R})
=(U^{d}_{1 :},...,U^{d}_{i_d:},...,U^{d}_{N_d:})^{\top}$, $d\in[D]\footnote{[N] denotes the set \{1,2,3,...,N\}.}$, and $R$ is the CP rank of tensor  $\mathcal{X}$.

With the $D$-way CP decomposition defined in Equation (\ref{equ-cp}), we can easily reformulate it into element-wise form,
\begin{equation}\label{equ-elementwiseCP}
\mathcal{X}(i_1,i_2,...,i_D)=\sum_{r=1}^R U^{1}_{i_1,r} U^{2}_{i_2,r} \cdot \cdot \cdot U^{D}_{i_D,r} \nonumber
\end{equation}
where $\mathcal{X}(i_1,i_2,...,i_D)$ is the entry of tensor $\mathcal{X}$ with index $(i_1,i_2,...,i_D)$.

For the observed tensor $\mathcal{Y}$ can be factorized as a tensor $\mathcal{X}$ with a noise term $\boldsymbol{\varepsilon}$,
\begin{eqnarray}\nonumber
  p(\mathcal{Y}|\mathcal{X}) &=& \mathcal{N}(\mathcal{Y}|\mathcal{X}, \boldsymbol{\varepsilon}) \\\nonumber
   &=& \prod_{i_1,...,i_D}\mathcal{N}(\sum_{r=1}^R \prod_{d=1}^{D} U^{d}_{i_d,r}, \varepsilon^{-1})\\ \nonumber
   &=& \prod_{i_1,...,i_D}\mathcal{N}(x, \varepsilon^{-1}) \nonumber
\end{eqnarray}
where $x:=\mathcal{X}(i_1,...,i_D)$ denotes the tensor entry of index $(i_1,...,i_D)$, and $x$ follows univariate Gaussian distribution with mean and variance denoted by $\mu$ and $\sigma^2$ respectively.

\subsection{Variational Auto-encoder CP}
\begin{figure}
  \centering
  \includegraphics[scale=0.28]{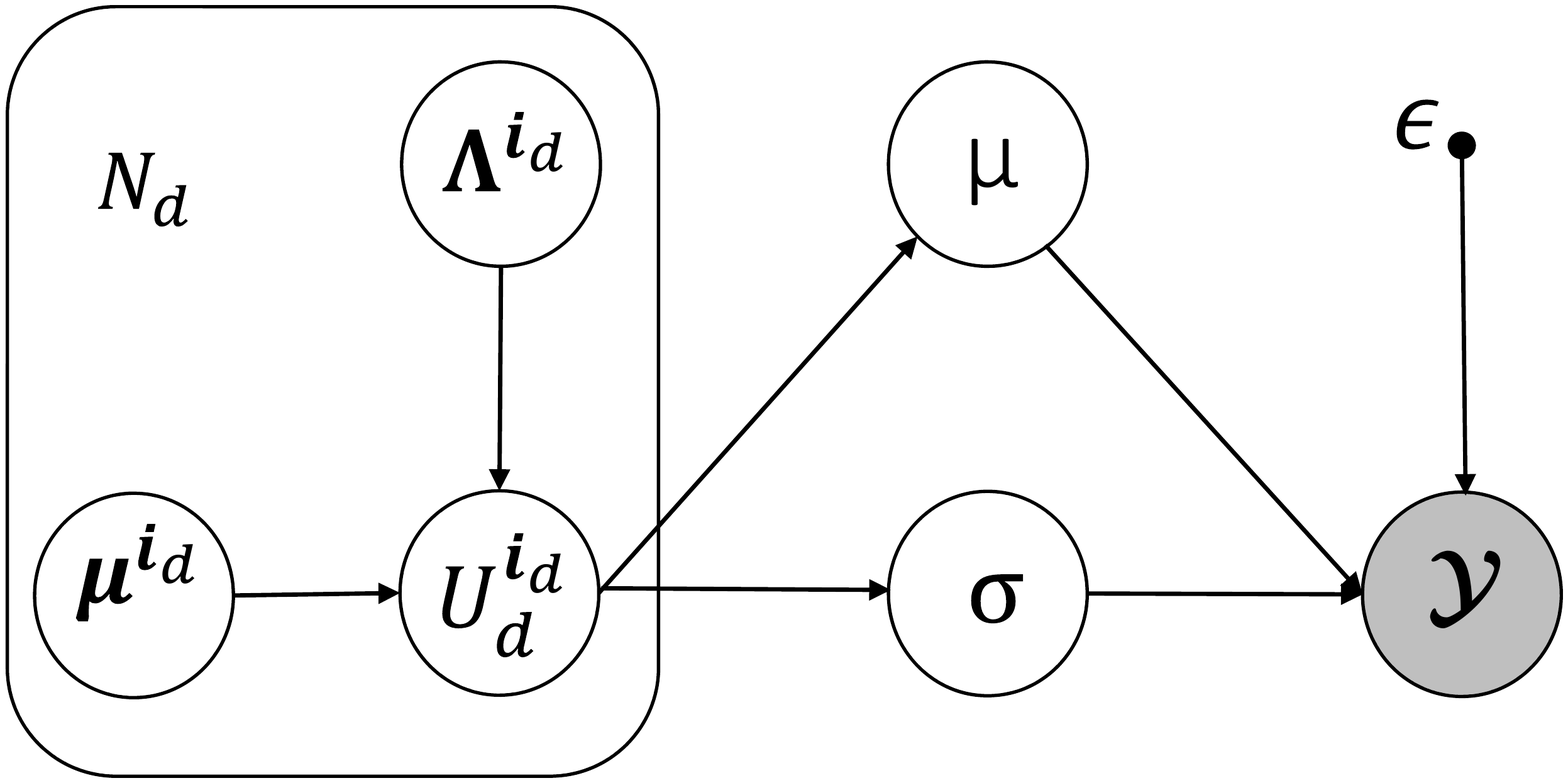}\\
  \caption{The graphical illustration of the VAECP.}\label{graphicalModel}
\end{figure}

The latent factors in Equation. (\ref{equ-cp}) can be drawn from a prior $p(U)$, and the likelihood conditioned on $U$ is $p(\mathcal{X} | U)$.
Contrary to traditional Bayesian model, for the $p(\mathcal{X} | U)$, we proposed that both $\mu$ and $\sigma^2$ are two functions of the latent CP factors as follows,
\begin{equation}
\nonumber
  p(\mathcal{X}|\{U^d\}_{d=1}^D) = \prod_{i_1,...,i_D}[\mathcal{N}(x|  \mu, \sigma^2)]^I\nonumber
\end{equation}
where $\mu:={\mu(U^{1}_{ i_1:}},..,{U^{D}_{i_D:}}), \sigma^2:=\sigma^2(U^{1}_{i_1 :},..,U^{D}_{i_D:})$, $U^{d}_{i_d:}$ is the $i_d$-th row of factor matrix $U^d$, and $I:=I(i_1,...,i_D)$ is a indicator function (equals to 1 if the $(i_1,...,i_D)$-th element is observed, 0 otherwise ).

We further consider the two functions $\mu$ and $\sigma^2$ can be represented as the outputs of two Neural Networks with the input of latent CP factors $U^{1}_{ i_1:},..,{U^{D}_{i_D:}}$ as follows,
\begin{alignat}{2}\nonumber
\mu=\boldsymbol{w}_{\mu}^{\top}\boldsymbol{h} + b_{\mu}\\  \nonumber
\log \sigma^2=\boldsymbol{w}_{\sigma}^{\top}\boldsymbol{h} + b_{\sigma} \nonumber
\end{alignat}
where $\boldsymbol{h}:=\boldsymbol{h}(U^{1}_{i_1 :},..,U^{D}_{i_D:})$ is the hidden layer shared by these two Neural Networks,

\begin{equation}\label{equ-neuralNet}
\boldsymbol{h}(U^{1}_{i_1 :},..,U^{D}_{i_D:}) = tanh(W^{\top}\mathbf{u}+\boldsymbol{b})
\end{equation}
and $\mathbf{u}=(U^{1}_{i_1 :};..;U^{D}_{i_D:})\in \mathbb{R}^{DR\times 1}$, $W \in \mathbb{R}^{D\times K}$ is a tensor, $W_{dk}\in \mathbb{R}^{R}$ are vectors. The parameters $\{W, \boldsymbol{w}_{\mu}, \boldsymbol{w}_{\sigma}\}$ and $\{\boldsymbol{b}, b_{\mu},b_{\sigma}\}$ are the weights and biases of the Neural Networks.
Then we have the posterior parameters $\Theta = \{W,\boldsymbol{w}_{\mu},\boldsymbol{w}_{\sigma}, b_{\mu},b_{\sigma}\}$. Figure \ref{graphicalModel} shows the graphical model of our method.

Then the goal of model inference is to calculate the posterior $p(U | \mathcal{X}) = p(\mathcal{X} | U, \Theta) p(U) / p(\mathcal{X})$. However, the calculation of the evidence $p(\mathcal{X})$ requires exponential time cost. Using variational inference, we can approximates this posterior with a family of distributions $q(U | \mathcal{X}, \Phi)$ involving the parameter $\Phi$.

In order to perform the variational inference within the framework of AEVB\citep{KingmaW13}, we further impose the approximated Gaussian posterior with the form as follows,

\begin{equation}\label{equ-priorLatentVariables}
q(U^{d}_{i_d:}| \boldsymbol \mu_{i_d}^d, \boldsymbol \Lambda_{i_d}^d) = \mathcal{N}(U^{d}_{i_d:}|\boldsymbol \mu_{i_d}^d,  (\boldsymbol \Lambda_{i_d}^d)^{-1}))
\end{equation}
where $( \boldsymbol \Lambda_{i_d}^d)^{-1}=diag(\boldsymbol \lambda^d_{i_d})$, $\boldsymbol \mu_{i_d}^d, \boldsymbol \lambda_{i_d}^d \in \mathbb{R}^{R\times 1}$. And for the given approximated posterior, the parameter $\Phi= \{\boldsymbol \mu_{i_d}^d,  \boldsymbol \lambda_{i_d}^d \}_{d=1}^D$.

In searching the optimal approximate posterior $q(U | \mathcal{X}, \Phi^*)$, we have to keep $q(U | \mathcal{X}, \Phi)$ close to the true posterior $p(U | \mathcal{X})$ with Kullback-Leibler divergence,

\begin{eqnarray}\label{eq:KL_posteriors} \nonumber
  q(U | \mathcal{X}, \Phi^*) &=& \mathop{\mathrm{argmin}}_{\Phi} KL(q(U | \mathcal{X}, \Phi) || p(U | \mathcal{X}) ) \\
   &=& \mathop{\mathrm{argmin}}_{\Phi}( \log p(\mathcal{X})-\mathcal{L}(\Phi|\mathcal{X}) ) \\ \nonumber
\end{eqnarray}
 where $\mathcal{L}(\Phi|\mathcal{X}) = \mathbb{E}_{q}[\log p(\mathcal{X}, U)] - \mathbb{E}_{q}[\log q(U | \mathcal{X}, \Phi)]$. Because Kullback-Leibler divergence is always greater than or equal to zero. So we get the lower bound $\mathcal{L}(\Phi|\mathcal{X})$ of $\log p(\mathcal{X}) $ from Equation. (\ref{eq:KL_posteriors}),
\begin{equation}\label{eq:lowerBound_PX}
\begin{aligned}\log p(\mathcal{X}) \geq & \mathcal{L}(\Theta, \Phi|\mathcal{X}) \\
  =&\mathbb{E}_{q}[\log p(\mathcal{X}|U,\Theta)]-KL(q(U | \mathcal{X}, \Phi)||p(U))
 &
 \end{aligned}
\end{equation}
It means that minimizing the Kullback-Leibler divergence is equivalent to maximizing the lower bound. And from Equation. (\ref{eq:lowerBound_PX}) we find that the lower bound involves both $\Phi$ and $\Theta$. Because there are no global representations of latent variables that are shared by all data points, we can get the lower bound of a single data point $x^{i}$,
\begin{equation}\label{eq:lowerBound_px}
\mathcal{L}(\Theta, \Phi|x^{i}) =\mathbb{E}_{q}[\log p(x^{i}|\mathbf{u}^{i},\Theta)] - KL(q(\mathbf{u}^{i}|x^{i},\Phi)||p(\mathbf{u}))
\end{equation}
where $\mathbf{u}=(U^{1}_{i_1 :};..;U^{D}_{i_D:})$.
In most scenario (with non-conjugate setting), the expectation term $\mathbb{E}_{q}[\log p(x^{i}|\mathbf{u}^{i},\Theta)]$ of Equation.~(\ref{eq:lowerBound_px}) is intractable. We cannot derive the gradient of lower bound w.r.t its parameters directly. Kingma et al. parameterized the latent variable $\mathbf{u}^{i} \sim q(\mathbf{u}^{i}|x^{i},\Phi)$ in the expectation term using a differentiable transformation $g_{\Phi}(\bm\epsilon)$ of an additional noise variable $\bm \epsilon$ \citep{KingmaW13}, then we have,
\begin{equation}\label{equ-reparameterizeForMean}
U^{d}_{i_d:} = \boldsymbol \mu_{i_d}^d + diag(({\boldsymbol \lambda}_{i_d}^d)^{-1/2})) \boldsymbol \epsilon_{i_d} \nonumber
\end{equation}

The first term $\mathbb{E}_{q}[\log p(x^{i}|\mathbf{u}^{i},\Phi)]$ of Equation. (\ref{eq:lowerBound_px}) is the reconstruction loss or expected negative log-likelihood of the data point $x^{i}$. The expectation is taken with respect to the encoder's distribution over the distribution of latent variable $\mathbf{u}$. This expected negative log-likelihood term encourages the decoder to try to reconstruct the data. Supposing  that the decoder's output does not reconstruct the data well, it will incur a large loss.
We regularize the reconstruction loss with the second term. It is the Kullback-Leibler divergence between the recognition model's distribution $q(\mathbf{u}^{i}|x^{i},\Phi)$ and $p(\mathbf{u})$. This divergence measures how close $q$ is to  our prior $p$. If the representations of output $u$ in recognition model that are different than those from the prior, there will be a penalty of the cost.

To model our problem within the architecture of Bayesian inference, we assign shared Gaussian priors over the latent variables $U^{d}_{: i_d}$ with the parameters $\Psi = \{\tilde{\boldsymbol \mu}, \tilde{\boldsymbol \Lambda}\}$,
\begin{equation}\label{equ-prior}
p(U^{d}_{i_d:}|\tilde{\boldsymbol \mu}, \tilde{\boldsymbol \Lambda}) = \mathcal{N}(U^{d}_{ i_d:}|\tilde{\boldsymbol \mu},  \tilde{\boldsymbol \Lambda}^{-1}) \nonumber
\end{equation}
where the mean and variance are $\tilde{\boldsymbol \mu}$ and $\tilde{\boldsymbol \Lambda}^{-1}$ respectively, and
$( \tilde{\boldsymbol\Lambda})^{-1}=diag(\tilde{\boldsymbol \lambda})$, and $\tilde{\boldsymbol \lambda} \in \mathbb{R}^{R\times 1}$.

The loss function (lower bound) of the decomposed form on every single data point is,
\begin{equation}\label{likelihood_lower_bound}
\begin{aligned} \mathcal{L}(\Theta,\Phi, \Psi |\mathcal{X})=
 &\sum_{l=1}^{L}\sum_{i_1=1}^{N_1}\cdot\cdot\cdot \sum_{i_D=1}^{N_D}\frac{I_{i_1,...,i_D}}{L} \log \mathcal{N}(x|\mu^{(l)}, {\sigma^2}^{(l)})\\
-&\sum _{d=1}^D \sum_{i_d=1}^{N_d}KL[q(U^{d}_{i_d:}|\boldsymbol\mu_{i_d}^d, \boldsymbol \lambda_{i_d}^d) || p(U^{d}_{i_d:}|\tilde{\boldsymbol \mu}_{i_d}^d, \tilde{\boldsymbol \Lambda}_{i_d}^d) ]
 \end{aligned}
\end{equation}
where $\boldsymbol \epsilon_{i_d}\sim \mathcal{N}(0,I)$, and lower bound related with three kinds of parameters $\Theta$, $\Phi$ and $\Psi$.
As we discussed before, the KL term of  Equation. (\ref{likelihood_lower_bound}) is a regularization over reconstruction loss (the first term). This term tries to keep the representations latent factors of each tensor entry as similar as the prior distribution.

\section{Optimization}
In the VAECP model, we assign independent and different posterior distributions for latent factors (as shown in Equation. (\ref{equ-priorLatentVariables}) ). Hence,
the distributions of the latent variables for each tensor entries are independent.
It means that our model can be trained with the stochastic gradient descent to optimize the loss with respect to the parameters of the encoder and decoder. In this paper, we choose a popular stochastic optimization method Adam \citep{adam2015ICLR}.

The optimization of Equation. (\ref{likelihood_lower_bound}) can be performed by randomly splitting the available observed elements of tensor $\mathcal{Y}$ into minibatches. The process of optimization with Adam present as shown in Algorithm \ref{algo_vaecp}  . with the parameters $\Omega = \Omega_0$, step size $\alpha$, $\beta_1=0.9, \beta_2=0.999 , \epsilon= 10^{-8}$ (note that $\beta_1$, $\beta_2$ and $\epsilon$ are fixed by Adam \citep{adam2015ICLR}) and minibatches of size 30.

\begin{algorithm}
\SetKwData{Left}{left}\SetKwData{This}{this}\SetKwData{Up}{up}
\SetKwFunction{Union}{Union}\SetKwFunction{FindCompress}{FindCompress}
\SetKwInOut{Input}{Input}\SetKwInOut{Output}{output}
\Input{Initial tensor data in minibatch $\mathcal{Y}_{batch}$, $K,\epsilon$}
\Output{Updated parameters $\Omega=\{\Theta, \Phi, \Psi\}$}
\BlankLine
{Initialize:$\Omega = \Omega_0$, step size $\alpha$, $\beta_1, \beta_2, \epsilon, m_0, v_0$, and step $t = 0$}\;
\For{$\mathcal{Y}_{batch}$}{
\While{$\Omega_t$ not converged}{
$loss=-\mathcal{L}(\Omega_{t-1} |\mathcal{Y}_{batch} )$\;
step $t = t +1 $ \;
$grad_{t} = \frac{\partial loss}{\partial \Omega_{t-1}}$\;
$m_t = \beta_1m_{t-1} + (1-\beta_1)grad_{t}$\;
$v_t = \beta_2v_{t-1} + (1-\beta_2)grad_{t}^2$\;
$\Omega_t = \Omega_{t-1} - \alpha \frac{m_t}{1-\beta_1^t} / (\sqrt{\frac{v_t}{1-\beta_2^t}} + \epsilon)$\;
}}
\caption{VAECP}\label{algo_vaecp}
\end{algorithm}
\section{Experiments}
We evaluate the performance of our proposed method in this section.

\subsection{Baselines}
Here, we compared our algorithm with several state-of-the-art methods: Nonnegative CP (NCP) \citep{shashua2005non}, High Order SVD (HOSVD) \citep{de2000multilinear}, Alternating Least-Squares CP  (CP) \citep{harshman1970foundations}, \citep{comon2009tensor, kolda2009tensor},  Alternating Least-Squares Tucker (Tucker) \citep{kapteyn1986approach, kolda2009tensor} and Bayesian CP factorization methods including \citep{zhao2015bayesian}(FBCP), Bayesian Tucker factorization method (InfTucker) \citep{xu2011infinite}).

\subsection{Evaluation Metric}
To evaluate the efficacy of our method and other compared algorithms, we use the root mean square error (RMSE) of the metric of prediction. Different from existing methods, in this paper, we just evaluate the RMSE on the test data, which is calculated as
$RMSE = \sqrt{\frac{1}{|\mathcal{X}_{u}|}\sum(\mathcal{X}_{u} - \mathcal{X}^{'}_{u})^2}$,
where $\mathcal{X}_{u}$ is the unobserved part of original tensor and $\mathcal{X}_{u}^{'}$ is the prediction, $|\mathcal{X}_{u}|$ is the cardinality of the unobserved set. We don't calculate RMSE on whole dataset because our model coupled to neural networks and it likely to cause overfitting on training data (RMSE on training may be 0). So we only computed the RMSE for testing samples of tensor.

\subsection{Datasets}
\textbf{Synthetic Data}. The size of synthetic data is $20 \times 20 \times 20$. To get it, we firstly generate three factor matrices of same sizes $20 \times r$, and each row of them is randomly drawn from $\mathcal{N}(\textbf{0},\textbf{I}_{r})$. Then we can get synthetic tensor with additional noise from $\mathcal{N}(0,1)$ and the CP $rank = r$.

\textbf{Real-world Datasets}.
We evaluate our models on three chemometrics datasets \footnote{http://www.models.life.ku.dk},  Amino Acid,  Flow Injection analysis\citep{xu2011infinite, morup2009automatic} and the Sugar Process data \citep{bro1999exploratory}. These data are widely used to evaluate the performance of tensor decomposition in a number of tensor literature such as \citep{xu2011infinite, morup2009automatic}.
The sizes of these datasets are $5 \times 51 \times 201, 12 \times 100 \times 89$ and $265 \times 571 \times 7$, respectively.

Amino Acid data \citep{xu2011infinite, morup2009automatic} consists five laboratory-made amino acid samples. This data set contains the excitation and emission spectra of five samples of different amounts of tyrosine, tryptophane and phenylalanine forming a $5(samples) \times 51(excitation) \times 201(emission)$ array.

Flow Injection analysis data \citep{xu2011infinite, morup2009automatic} is given by the absorption spectra over time for three different chemical analytes measured in 12 samples with different concentrations, i.e. $12(samples) \times 100(wavelengths) \times 89(times)$ \citep{bro1997parafac}.

For every sample of Sugar process data \citep{bro1999exploratory}, the emission spectra from 275-560 nm were measured in 0.5 nm intervals (571 wavelengths) at seven excitation wavelengths (230, 240, 255, 290, 305, 325, 340 nm). The data of all the 265 samples can be arranged in an $i \times j \times k$ three-way array of specific size $265 \times 571 \times 7$. The first mode refers to samples, the second to emission wavelengths, and the third to excitation wavelengths. The $(i, j, k)$th element in this array corresponds to the measured emission intensity from sample $i$, excited at wavelength $k$, and measured at wavelength $j$.

All the data points of the above data were normalized. The entries of the tensors have the mean of 0 and unit variance. For training, we use 5-fold cross validation for the training data. We repeated this process in 10 times.

\subsection{Evaluation on the Synthetic Dataset}
\textbf{Experimental setting}. In this experiment, we test our algorithm on a synthetic tensor.
The learning rate $\alpha$ (Algorithm. \ref{algo_vaecp}) was set as 0.0001 in this test.
For the synthetic data, the CP rank is already known ( rank=$r$). The compared methods need to make estimations of rank as the inputs. We evaluate RMSE of them by varying the rank $r$ in the candidate set $\{9,10,11,12\}$.  For our algorithm,  there are two important parameters of the Neural Networks $R$ and $K$ as shown in Equation. (\ref{equ-neuralNet}).
$R$ is the `rank' corresponding to the size of latent factors $U^{d}_{i_d:}$. Therefore, we set $R$ that equals to $r$ from the rank set $\{9,10,11,12\}$. We set $K=50$ for the synthetic test.
We compared the prediction error of all the approaches with 80\% of the data are held out for training and used the remaining 20\% for testing in each run.

\begin{figure}
  \centering
  \includegraphics[scale=0.55]{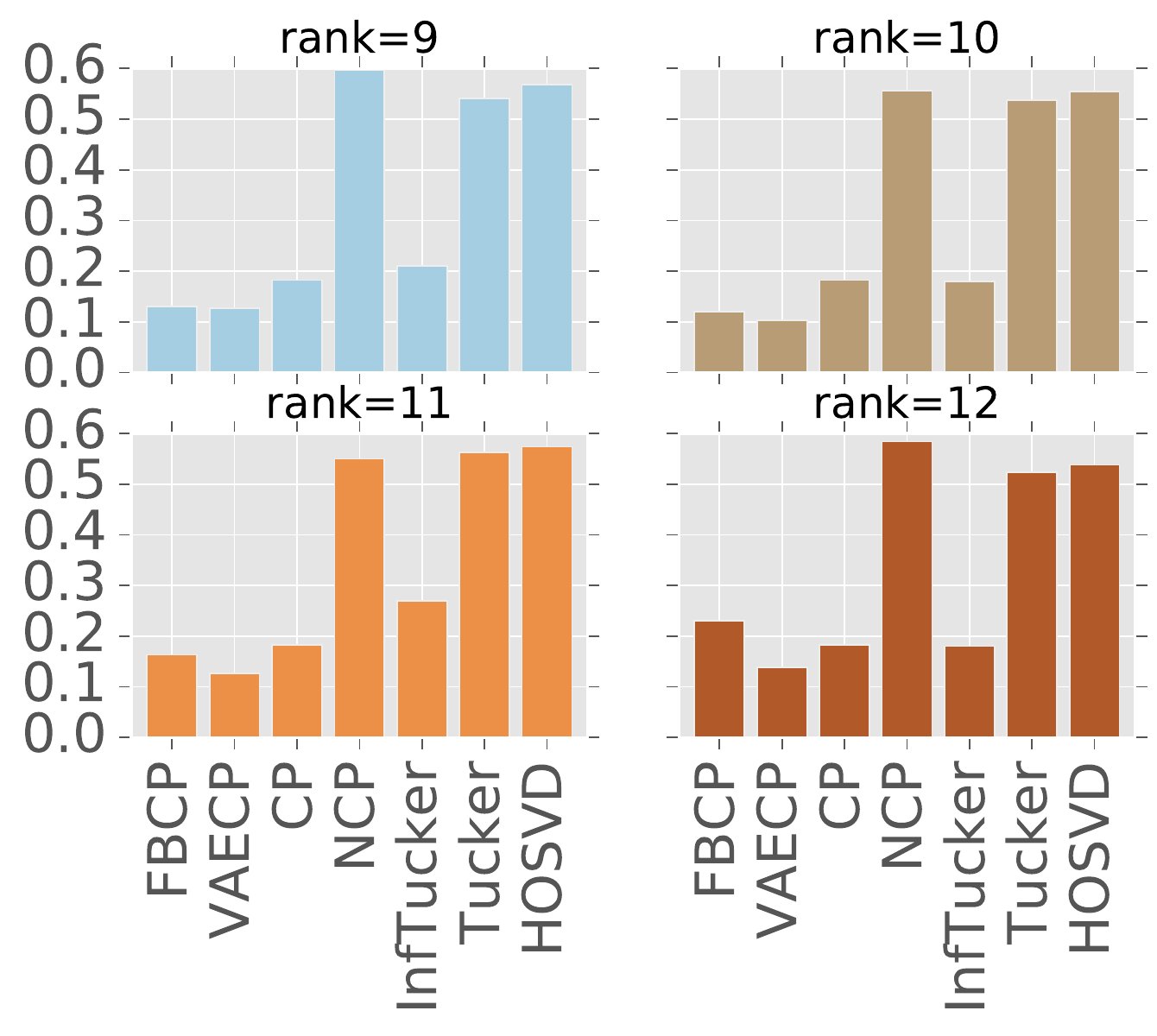}\\
  \caption{Performance of testing VAECP and other state-of-the-art algorithms on Synthetic dataset. Y-axis is the RMSE. }\label{syn_rmse}
\end{figure}

\textbf{Results}.
The results present in Figure \ref{syn_rmse}. We find that Bayesian methods performance better than traditional approaches. And our model achieves lowest prediction error than the traditional CP methods.

\subsection{Evaluation on Real Datasets}

\textbf{Experimental setting}.   For the real datasets, we do not know the true rank. However, most of algorithms,  NCP, HOSVD, CP, Tucker,  work with a guess of rank as input. FBCP can determine a rank automatically. In VAECP, the conception of `rank' is not so clear because it is related with the parameters of Neural Networks. In our experiments, we feed other algorithms with a rank that FBCP think it is the best. Although it is a little bit unfair for FBCP, there is no better way to compared all of them together. Of course, we will study the performance of VAECP by varying the `rank' $R$. The `rank' we discussed here is not exactly as same as it defined in CP decomposition. In order to study $R$, we fix $K=100$ for VAECP in all the following experiments (we actually chose different $K$ (50 for synthetic, 100 for real) according to the sizes of datasets.).

\begin{figure}[H]
  \centering
  \includegraphics[scale=0.55]{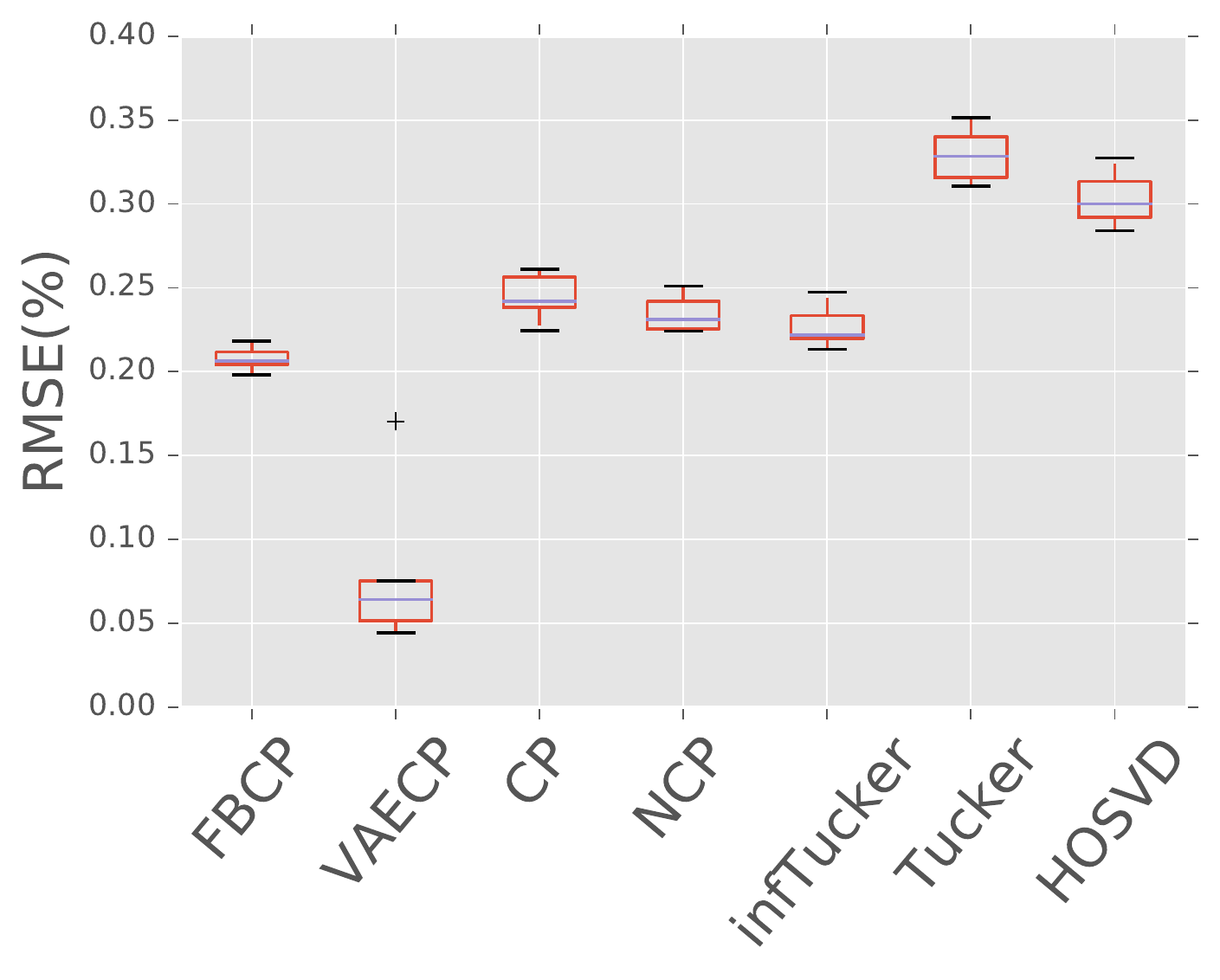}\\
  \caption{Performance of testing VAECP and other state-of-the-art algorithms on the Flow Injection analysis dataset. }\label{fia_rmse}
\end{figure}

\begin{figure}[H]
  \centering
  \includegraphics[scale=0.55]{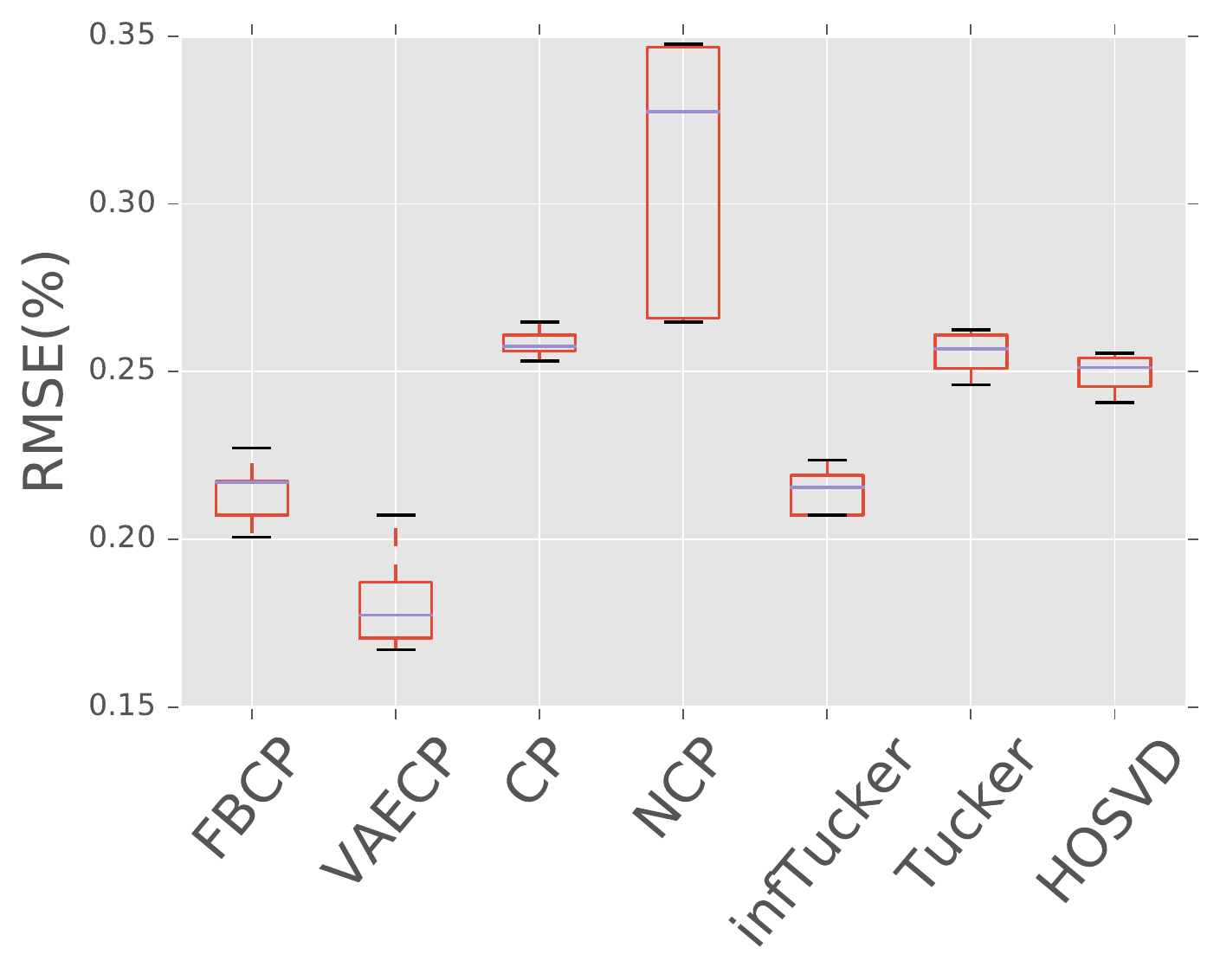}\\
  \caption{Performance of testing VAECP and other state-of-the-art algorithms on the Sugar process dataset. }\label{sugar_rmse}
\end{figure}

We also randomly get 80\% of all the data points for training, and the rest for testing. The training strategy is also the 5-fold cross validation we mentioned before.
The recommended learning rate $\alpha$ can choose from $10^{-5}-5*10^{-5}$ .
\begin{figure}[H]
  \centering
  \includegraphics[scale=0.55]{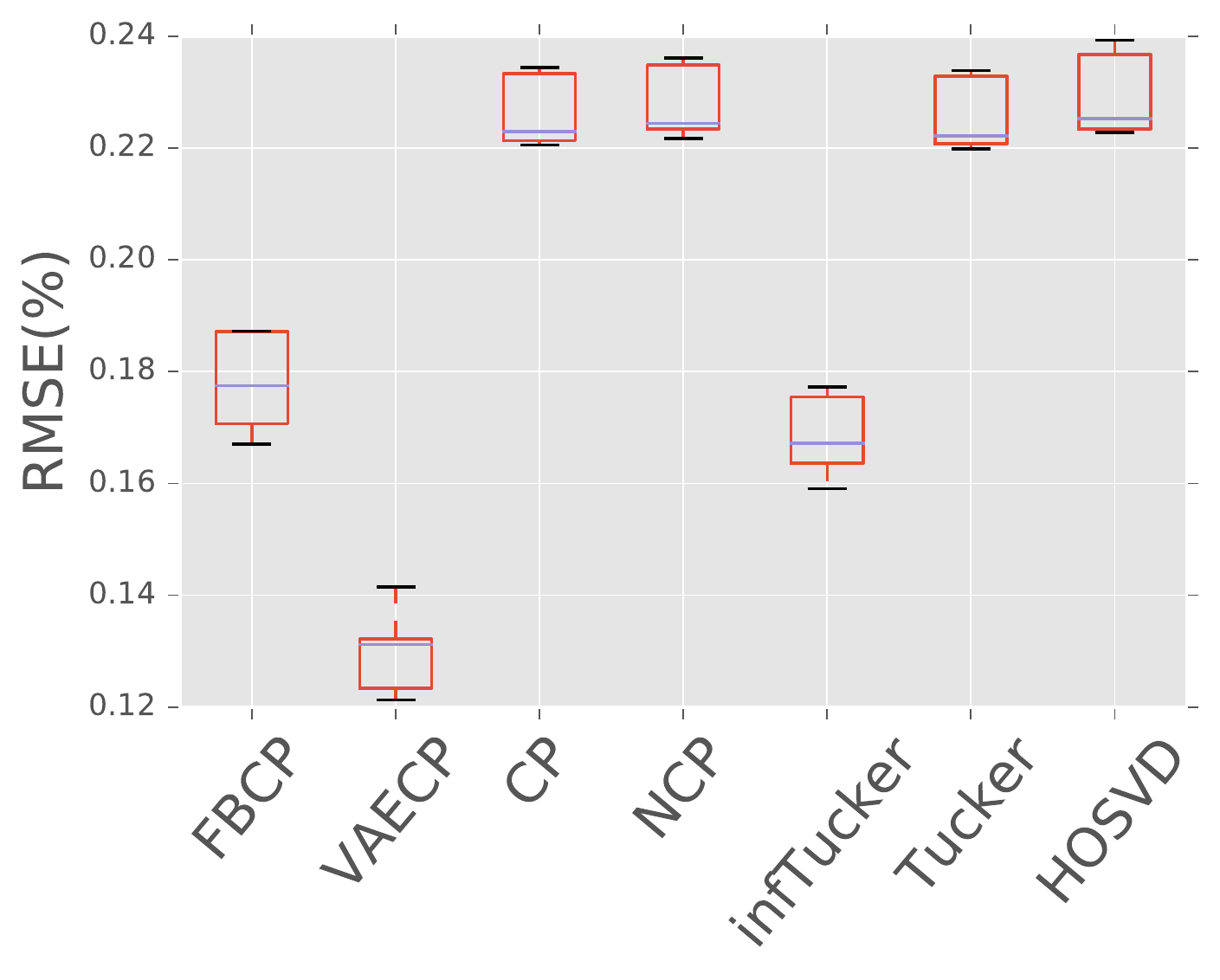}\\
  \caption{Performance of testing VAECP and other state-of-the-art algorithms on the Amino dataset.  }\label{amino_rmse}
\end{figure}
\textbf{Results}.   The run process was repeated as the way we discussed before. All the results from each run were recorded. We present all of them for VAECP and other approaches with \textit{boxplot} ( Figure \ref{amino_rmse} - \ref{sugar_rmse} ). The height of the box is the likely range of RMSE variation (distance between the RMSE of first quartile and the third quartile). The blue line in the box is the median value of RMSE. And the two black bars on the top and bottom of the box represent the maximum and minimum values. The ``+'' denotes the outlier marker.

Figure \ref{amino_rmse} reports the performance of VAECP and other methods on Amino Acid data. Bayesian-based method FBCP,  VAECP and infTucker perform better than multi-linear methods.
And our method achieves predict error better than all the gold standard CP decomposition algorithms.

The test on Flow Injection analysis data (Figure \ref{fia_rmse} ) also show that VAECP performs a better prediction than other compared methods. VAECP outperforms the other methods. However, we can clear find that there is an outlier of VAECP's performance above the box. The outlier tells us some stories. One possible explanation is that in one of the running, the training of the two layers Neural Networks might underfitting. The more complicated high-order relationships of Flow Injection analysis may cause this phenomenon. We may need to study this problem by putting a deeper network in our model as a future work.
Figure \ref{sugar_rmse} reports the RMSE of the prediction on Sugar process data. We also find that VAECP did a better prediction than other compared methods.

\subsection{Study the Tensor Rank of VAECP}
\begin{figure}
  \centering
  \includegraphics[scale=0.55]{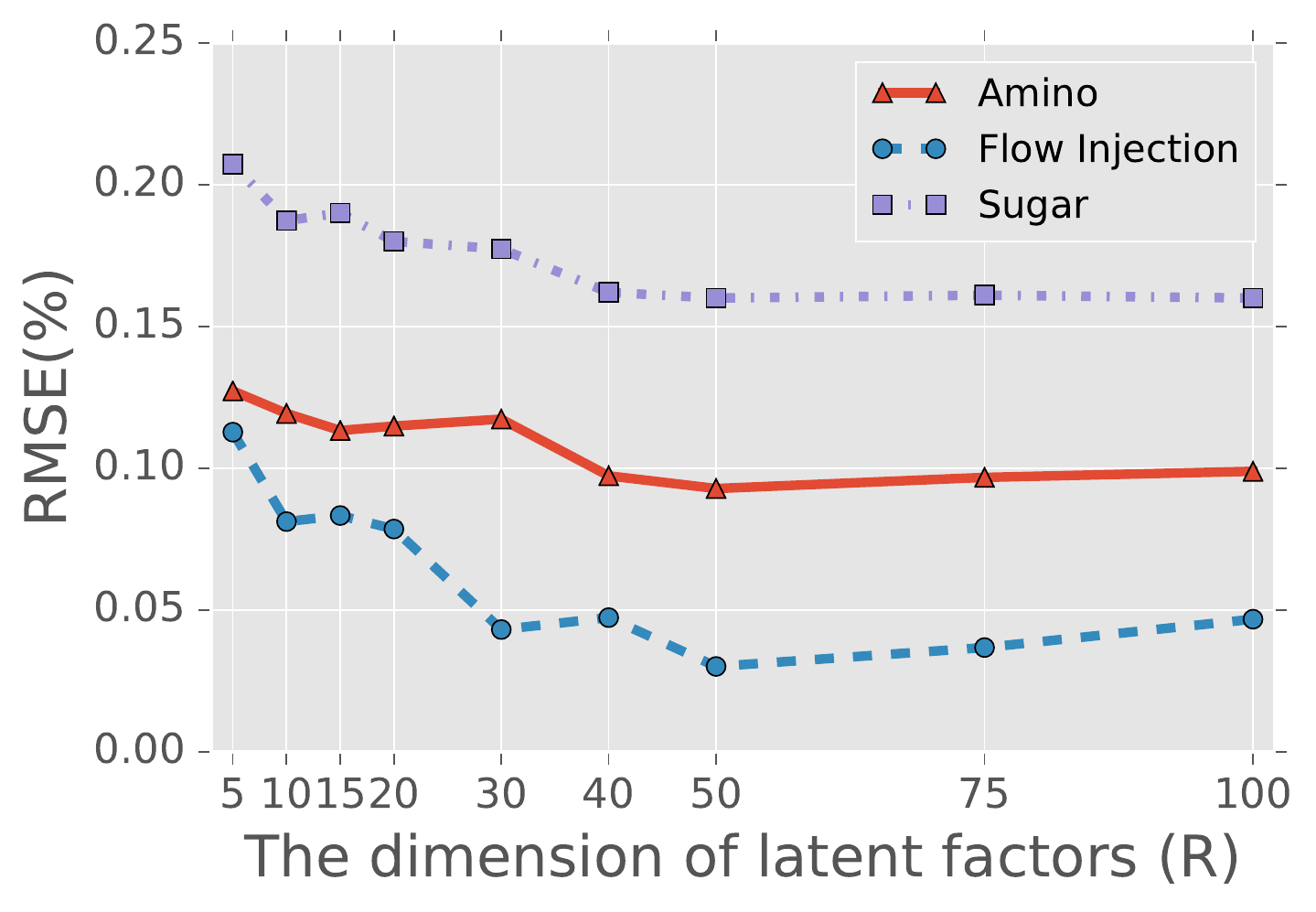}\\
  \caption{The RMSE values VAECP on three real world datasets. The dimensions of latent factors $R=5,10,15,20,30,40,50,75,100$.}\label{rmse_latent_dim}
\end{figure}

During the experiments, we found that the conception of tensor rank is not as clear as we defined in traditional tensor decomposition algorithms (for example, the CP rank). According to Equation. (\ref{equ-cp}) and (\ref{equ-neuralNet}), the `rank' of VAECP is $R$ (the size of latent factors $U^{d}_{i_d:}$). Figure \ref{rmse_latent_dim} illustrates the RMSE by varying the latent factors (the `rank') on three real-world datasets. It seems that the best `rank' of the three tests are all pretty high. By increasing the rank to 100, all the three datasets did not report a sharp drop of performance. Especially for the Sugar process data, the performance still can be improved if we assign an even larger `rank'.

\section{Conclusion}
In the view of Bayes, high order tensor can be generated from a complicated random process involving the latent factors. To model the relationship between latent factors and observed data, we propose to abstract this process with Neural Networks. The input of the Neural Networks are the latent factors and the output are the parameters of distributions for predicting the missing values. 
The experimental results on synthetic and three real-world tensor datasets show that our new model achieved significantly higher prediction performance than the most state-of-the-art tensor decomposition approaches.
One interesting thing we found is that the conception of tensor rank is not as clear as we defined in traditional tensor decomposition algorithms.  Even we input a `rank' that is out of the valid range defined by traditional methods, the performance of our approach will not fall sharply.
Our model proposed in this paper only focus on continuous tensor data, and the binary data analysis was left as a future work.

\bibliographystyle{aaai}
\bibliography{ref}

\newpage
\begin{appendices}
\section{Derivative Results for Gradients}
The derivative results,

For $\Theta=\{W, \boldsymbol{b}, \boldsymbol{w}_{\mu}, \boldsymbol{w}_{\sigma}, b_{\mu}, b_{\sigma}\}$, where $W\in \mathbb{R}^{K\times DR}, \boldsymbol{w}_{\mu}, \boldsymbol{w}_{\sigma} \in \mathbb{R}^{K\times 1},  b_{\mu}, b_{\sigma} \in \mathbb{R}$.

The derivative for $\Theta$,

\begin{equation}\label{equ-derivative_w_mu}
\frac{\partial L}{\partial \boldsymbol{w}_{\mu} } = \sum_{l=1}^{L} \sum_{i_1=1}^{N_1}\cdot\cdot\cdot \sum_{i_D=1}^{N_D} \frac{I(i_1,...,i_D)}{L} \sigma^{-2}(x-\mu)\boldsymbol h  \nonumber
\end{equation}

\begin{equation}\label{equ-derivative_b_mu}
\frac{\partial L}{\partial b_{\mu} } = \sum_{l=1}^{L} \sum_{i_1=1}^{N_1}\cdot\cdot\cdot \sum_{i_D=1}^{N_D} \frac{I(i_1,...,i_D)}{L} \sigma^{-2}(x-\mu)  \nonumber
\end{equation}

\begin{equation}\label{equ-derivative_w_sigma}
\frac{\partial L}{\partial \boldsymbol{w}_{\mu} } = \sum_{l=1}^{L} \sum_{i_1=1}^{N_1}\cdot\cdot\cdot \sum_{i_D=1}^{N_D} \frac{I(i_1,...,i_D)}{L} (\frac{(x-\mu)^2}{2\sigma^2}-1/2)\boldsymbol h  \nonumber
\end{equation}

\begin{equation}\label{equ-derivative_b_sigma}
\frac{\partial L}{\partial b_{\sigma} } = \sum_{l=1}^{L} \sum_{i_1=1}^{N_1}\cdot\cdot\cdot \sum_{i_D=1}^{N_D} \frac{I(i_1,...,i_D)}{L} (\frac{(x-\mu)^2}{2\sigma^2}-1/2)  \nonumber
\end{equation}

\begin{eqnarray}\label{equ-derivative_W_dk}\nonumber
\frac{\partial L}{\partial W_{dk} } &=& \sum_{l=1}^{L}  \frac{I(i_1,...,i_D)}{L} [
\frac{x-\mu}{\sigma^2} \frac{\partial \mu}{\partial W_{dk} } \\\nonumber
&+& \frac{(x-\mu)^2}{\sigma^3} \frac{\partial \sigma}{\partial W_{dk}} - \frac{1}{\sigma} \frac{\partial \sigma}{\partial W_{dk}}\\
&=& [\frac{1}{2}(\frac{(x-\mu)^2}{\sigma ^2} - 1)\boldsymbol{w}_{\sigma}^k + \frac{(x-\mu)}{\sigma ^2}\boldsymbol{w}_{\mu}^k] tanh' U^{d}_{i_d:} \nonumber
\end{eqnarray}

\begin{equation}\label{equ-derivative_B}\nonumber
\frac{\partial L}{\partial \boldsymbol{b} }= [\frac{1}{2}(\frac{(x-\mu)^2}{\sigma ^2} - 1)\boldsymbol{w}_{\sigma} + \frac{(x-\mu)}{\sigma ^2}\boldsymbol{w}_{\mu}] \nonumber
\end{equation}
where

\begin{eqnarray}\nonumber
   \frac{\partial \mu}{\partial W_{dk}}&=& \boldsymbol{w}_{\mu}^k tanh' U^{d}_{i_d:}\\
  \frac{\partial \sigma}{\partial W_{dk} } &=& \frac{\sigma \boldsymbol{w}_{\sigma}^k}{2} tanh' U^{d}_{i_d:}   \nonumber
\end{eqnarray}

where $tanh' := tanh'(W_{dk}^{\top}U^{d}_{i_d:}) = 1-tanh^2 (W_{dk}^{\top}U^{d}_{i_d:})$

where $\sigma:=\sigma^2({U^{1}_{i_1:}}^{(l)},..,{U^{D}_{i_D:}}^{(l)})$,$x:=\mathcal{X}(i_1,...,i_D)$ ,$\mu:={\mu(U^{1}_{i_1:}}^{(l)},..,{U^{D}_{i_D:}}^{(l)})$ and $\boldsymbol{h}:=\boldsymbol{h}
(U^{1}_{i_1:},..,U^{D}_{i_D:})$

Derivative for $\Psi=\{\tilde{\boldsymbol \mu}, \tilde{\boldsymbol \lambda}\}$
\begin{eqnarray}\label{equ-derivative_mu_psi}  \nonumber
\frac{\partial L}{\partial \tilde{\boldsymbol \mu} } &=& \tilde{\boldsymbol \lambda} \odot (\boldsymbol{\mu}_{i_d}^d - \tilde{\boldsymbol \mu})
\end{eqnarray}

\begin{eqnarray}\label{equ-derivative_lambda_psi}\nonumber
\frac{\partial L}{\partial \tilde{\boldsymbol \lambda} } &=& -\frac{1}{2} (\boldsymbol{\mu}_{i_d}^d - \tilde{\boldsymbol \mu})^2 + \frac{1}{2} (\frac{1}{\tilde{\boldsymbol \lambda}}- \frac{1}{\boldsymbol{\lambda}_{i_d}^d})
\end{eqnarray}

Derivative for $\Phi=\{\boldsymbol \mu_{i_d}^d, \boldsymbol \Lambda^{i_d}\}$
\begin{eqnarray}\label{equ-derivative_mu} \nonumber
\frac{\partial L}{\partial \boldsymbol{\mu}_{i_d}^d } &=& \tilde{\boldsymbol \lambda} \odot (\tilde{\boldsymbol \mu} - \boldsymbol{\mu}_{i_d}^d) \\  \nonumber
&+& \sum_{l=1}^{L}\frac{I(i_1,...,i_D)}{L} (-\frac{1}{2}\frac{\partial L_4}{\partial \boldsymbol{\mu}_{i_d}^d}- \frac{1}{\sigma} \frac{\partial \sigma}{\partial \boldsymbol{\mu}_{i_d}^d} )\\ \nonumber
&=& \tilde{\boldsymbol \lambda} \odot (\tilde{\boldsymbol \mu} - \boldsymbol{\mu}_{i_d}^d) + \sum_{l=1}^{L}\frac{I(i_1,...,i_D)}{L}(\sum_{k=1}^{K}m(k)tanh'  W_{dk})\\  \nonumber
\end{eqnarray}

where $L_4=\frac{(x-\mu)^2}{\sigma ^2}$, $m(k)=\frac{1}{2}(\frac{(x-\mu)^2}{\sigma ^2} - 1)\boldsymbol{w}_{\sigma}^k + \frac{2(x-\mu)}{\sigma ^2}\boldsymbol{w}_{\mu}^k$ and

\begin{eqnarray}  \nonumber
   \frac{\partial L_4}{\partial \boldsymbol{\mu}_{i_d}^d}&=& -2\sigma^{-2}(x-\mu)\sum_{k=1}^{K} \boldsymbol{w}_{\mu}^k tanh'  W_{dk}\\ \nonumber
   &-& L_4 \boldsymbol{w}_{\sigma}^k tanh'  W_{dk}\\ \nonumber
  \frac{\partial \sigma}{\partial \boldsymbol{\mu}_{i_d}^d } &=& \frac{\sigma }{2}\sum_{k=1}^{K}\boldsymbol{w}_{\sigma}^k tanh'  W_{dk} \nonumber
\end{eqnarray}
where $tanh':= (1-tanh^2(\sum_{d=1}^{D}W_{dk}^{\top}U^{d}_{i_d:}))$

\begin{eqnarray}\label{equ-derivative_lambda}\nonumber
\frac{\partial L}{\partial \boldsymbol{\lambda}_{i_d}^d } &=&  \frac{1}{2}  (\tilde{\boldsymbol \lambda}  (\boldsymbol{\lambda}_{i_d}^d)^{-2} - \frac{1}{\boldsymbol{\lambda}_{i_d}^d}) \\\nonumber
&+& \sum_{l=1}^{L}\frac{I(i_1,...,i_D)}{L} (-\frac{1}{2}\frac{\partial L_4}{\partial \boldsymbol{\lambda}_{i_d}^d}- \frac{1}{\sigma} \frac{\partial \sigma}{\partial \boldsymbol{\lambda}_{i_d}^d} )\\ \nonumber
&=& \frac{1}{2} (-\frac{1}{\boldsymbol{\lambda}_{i_d}^d} + (\boldsymbol{\lambda}_{i_d}^d)^{-2} ) \\\nonumber
&+& \sum_{l=1}^{L}\frac{I(i_1,...,i_D)}{L}(\sum_{k=1}^{K}n(k)tanh'  ({\boldsymbol \lambda}_{i_d}^d)^{-3/2}\odot  \boldsymbol \epsilon_{i_d} \odot W_{dk}) \nonumber
\end{eqnarray}

where $n(k)=\frac{1}{4}(1-\frac{(x-\mu)^2}{\sigma ^2} )\boldsymbol{w}_{\sigma}^k - \frac{(x-\mu)}{2\sigma ^2}\boldsymbol{w}_{\mu}^k$,

\begin{eqnarray}\nonumber
   \frac{\partial L_4}{\partial \boldsymbol{\lambda}_{i_d}^d}&=& -2\sigma^{-2}(x-\mu)\sum_{k=1}^{K} \boldsymbol{w}_{\mu}^k tanh'  W_{dk}\\\nonumber
   &-& L_4 \boldsymbol{w}_{\sigma}^k tanh'  W_{dk}\\\nonumber
  \frac{\partial \sigma}{\partial \boldsymbol{\lambda}_{i_d}^d } &=& -\frac{\sigma }{4}\sum_{k=1}^{K}\boldsymbol{w}_{\sigma}^k tanh' ({\boldsymbol \lambda}_{i_d}^d)^{-3/2}\odot  \boldsymbol \epsilon_{i_d} \odot W_{dk}\nonumber
\end{eqnarray}
where $\odot$ indicates the element-wise multiplication .

\end{appendices}
\end{document}